\documentclass{article}
\usepackage{amssymb}

\usepackage[final]{corl_2025} % Uncomment for the camera-ready ``final'' version.
\usepackage{enumitem}
\usepackage{amsmath}
\usepackage{graphicx}
\usepackage{booktabs}       % professional-quality tables
\usepackage{multirow}
\usepackage{wrapfig}

\title{ParticleFormer: A 3D Point Cloud World Model for Multi-Object, Multi-Material Robotic Manipulation}

\author{
  Suning Huang{$\dagger$}\\
  Stanford University\\ 
  \texttt{suning@stanford.edu}
  \And
  Qianzhong Chen\\
  Stanford University\\
  \texttt{qchen23@stanford.edu}
  \AND
  Xiaohan Zhang\\
  RAI Institute\\
  \texttt{xzhang@rai-inst.com}
  \And
  Jiankai Sun{$\ddagger$}\\
  Stanford University \\
  \texttt{jksun@stanford.edu}
  \And
  Mac Schwager{$\dagger$}\\
  Stanford University\\
  \texttt{schwager@stanford.edu}
}

\begin{document}
\def\thefootnote{$\dagger$}\footnotetext{Corresponding author}
\def\thefootnote{$\ddagger$}\footnotetext{Student advisor}
\maketitle

\newcommand{\model}{\texttt{ParticleFormer}}

%%%%%%%%%%%%%%%%%%%%%%%%%%%%%%%%%%%%%%%%%%%%%%%%%
\begin{abstract}
3D world models~(i.e., learning-based 3D dynamics models) offer a promising approach to generalizable robotic manipulation by capturing the underlying physics of environment evolution conditioned on robot actions. However, existing 3D world models are primarily limited to single-material dynamics using a particle-based Graph Neural Network model, and often require time-consuming 3D scene reconstruction to obtain 3D particle tracks for training. In this work, we present {\model}, a Transformer-based point cloud world model trained with a hybrid point cloud reconstruction loss, supervising both global and local dynamics features in multi-material, multi-object robot interactions. {\model} captures fine-grained multi-object interactions between rigid, deformable, and flexible materials, trained directly from real-world robot perception data without an elaborate scene reconstruction. We demonstrate the model's effectiveness both in 3D scene forecasting tasks, and in downstream manipulation tasks using a Model Predictive Control (MPC) policy.  In addition, we extend existing dynamics learning benchmarks to include diverse multi-material, multi-object interaction scenarios. We validate our method on six simulation and three real-world experiments, where it consistently outperforms leading baselines by achieving superior dynamics prediction accuracy and less rollout error in downstream visuomotor tasks. Experimental videos are available at \href{https://suninghuang19.github.io/particleformer_page/}{particleformer}.

\end{abstract}

\keywords{Robotic Manipulation, Learning-based Dynamics Modeling, Model-based Planning and Control} 

\vspace{-0.2em}
\section{Introduction}
\label{intro}
\vspace{-0.5em}

Recent advances in reinforcement learning (RL)~\citep{hafner2023mastering,huang2024dittogym,fu2024humanplus,peng2021amp,huang2024mentor} and imitation learning (IL)~\citep{lee2024behavior,zhao2023learning,chi2023diffusion,fu2024mobile,kim2024openvla} have enabled robots to acquire sophisticated manipulation skills across diverse tasks. However, most existing approaches directly map observations to actions, inherently coupling environmental dynamics with task-specific objectives. This tight coupling hinders generalization, as agents must learn solutions for every task-environment combination~\citep{zhou2024dino,etukuru2024robot,brohan2022rt}. In contrast, humans generalize effectively by forming flexible mental models of physical interactions and intuitively decoupling task goals from environment dynamics through visual perception~\citep{baillargeon1985object,battaglia2013simulation,watzl2017structuring}. Inspired by this insight, developing world models~\citep{ha2018world} that capture underlying physical interaction dynamics is crucial for improving generalization in robotic manipulation.

\begin{figure}[ht]
\centering
\includegraphics[width=1\linewidth, trim= 0cm 6.5cm 0cm 0cm, clip]{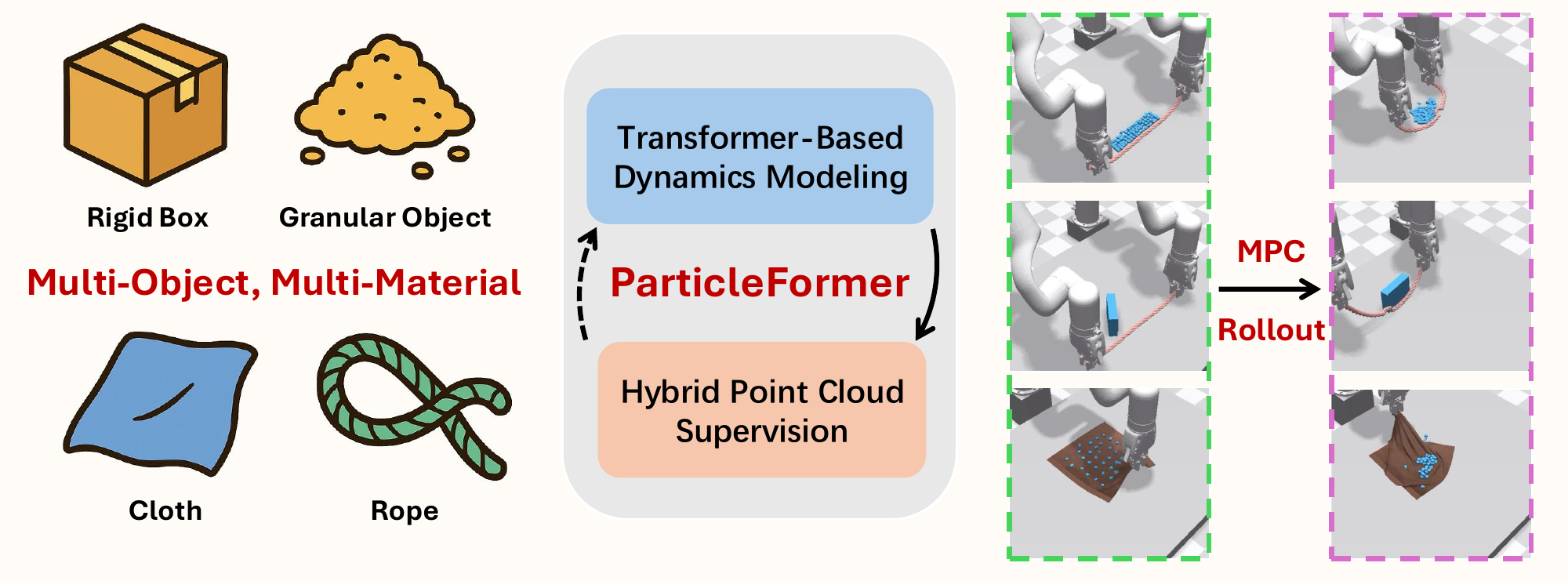}
\caption{\textbf{Motivation.} Modeling dynamics in multi-object, multi-material scenarios is challenging due to complex and heterogeneous interactions. In this paper, we propose {\model}, a Transformer-based point cloud world model trained with hybrid supervision, enabling accurate prediction and model-based control in robotic manipulation tasks.}
\label{fig:teaser}
\end{figure}

These world models have a long history in robotics and control~\citep{sutton1991dyna,todorov2005generalized}. Such models have been proposed with 2D (pixel space) or 3D (metric space) representations. For 3D world models, researchers have shown that by parameterizing objects as compositions of basic particles in 3D space and applying a graph-based neural dynamics~(GBND) framework to learn the interaction laws among these particles, the model can effectively represent scene dynamics for downstream planning and control~\citep{battaglia2018relational,zhang2024dynamic,sun2021adversarial,zhang2024adaptigraph,xue20233d,li2018learning,sun2022plate}. However, we observe that these methods constrain particle interaction learning to the topology defined by the graph, making the model sensitive to hyperparameters such as the maximum distance for establishing particle adjacency and the maximum number of neighbors. This sensitivity becomes particularly problematic in environments involving multi-material, multi-object interactions, which feature heterogeneous and complex contact dynamics. These models are also difficult to train with real-world data, often requiring time-consuming dynamic Gaussian Splatting (GSplat)~\citep{kerbl20233d,zhang2024dynamic} reconstruction to obtain 3D particle tracks, which are then used to supervise training of the particle dynamics model through an MSE loss. 2D world models bypass these issues by directly learning task-agnostic representations based on images, capturing dynamics within unified compact latent embeddings directly from pixel space~\citep{zhou2024dino,li2022neural,brooks2024video,hafner2023mastering,babaeizadeh2017stochastic}. However, these methods typically encode the 2D appearance of the scene into a single latent vector, which can fail to capture  underlying 3D environmental structure, limiting their applicability to compositional task scenes and fine-grained dynamics modeling.

To address these limitations, we present \textbf{ParticleFormer}, a Transformer-based~\citep{vaswani2017attention} 3D point cloud world model for multi-material robot manipulation, which achieves state-of-the-art performance in dynamics prediction and downstream visuomotor control tasks. Specifically, {\model}: (1) introduces a Transformer-based neural dynamics backbone that enables the model to automatically infer detailed 3D particle interactions among environment particles without manual hyperparameter tuning; and (2) leverages a hybrid point cloud loss that jointly optimizes the Chamfer Distance and a differentiable approximation of the Hausdorff Distance~\citep{charpiat2005approximations} to capture both local accuracy and global structural consistency.  Crucially, the loss is applied directly to point cloud sensor streams thereby avoiding costly dynamic GSplat scene reconstruction, while providing fine-grained supervision for accurate and robust dynamics modeling, even in multi-material, multi-object scenarios. Our simple yet effective framework consistently outperforms existing baselines across both synthetic and real-world benchmarks. Owing to the high quality of the dynamics predictions produced by {\model}, we can directly apply model predictive control (MPC)~\citep{kouvaritakis2016model,williams2017model} to achieve goal-directed behavior during evaluation.

Our key contributions are threefold. (i) First, we introduce a Transformer-based neural dynamics backbone to forecast the motion of point clouds representing the scene geometry. This improves both the accuracy and robustness of the dynamics prediction process. (ii) Second, we supervise the Transformer world model with Chamfer distance and Hausdorff distance loss terms applied directly to RGB sensor streams. This loss jointly captures local geometric accuracy and global structural consistency without requiring dense correspondence or explicit 3D reconstruction. (iii) Finally, we extend the current 3D dynamics modeling benchmarks to include multi-material, multi-object robot interactions, broadening the applicability to more diverse and practical scenarios. {\model} is experimentally evaluated on six simulation tasks and three real-world experiments, where results on both dynamics prediction accuracy and downstream visuomotor control demonstrate that our method consistently outperforms competitive baselines across all settings.

% Additionally, to achieve accurate 3D supervision, the above methods typically rely on multi-view synthesis using Neural Radiance Fields~\citep{mildenhall2021nerf} or Gaussian Splatting~\citep{kerbl20233d}, which can be inefficient and introduce additional computational burden during data collection, thereby complicating the overall pipeline.
%, where point clouds can be acquired from simple stereo images, thereby freeing the dynamics modeling pipeline from complicated 3D reconstruction.

\vspace{-0.5em}
\vspace{-0.2em}
\section{Related work}
\label{related_work}
\vspace{-0.5em}

\paragraph{World Model for Robotic Manipulation.}
Dynamics models are increasingly essential in robotics for predicting environment evolution from current states and actions~\citep{ha2018world,suh2021surprising,mitrano2021learning,lin2022learning,huang2022mesh}. Broadly speaking, there are two main lines of work:

\vspace{-0.3em}

\begin{itemize}[leftmargin=*,topsep=0pt]
    \item \textbf{Object-centric Modeling:} This approach represents objects as particles or meshes and uses GNNs to model local interactions among neighboring units~\citep{zhang2024adaptigraph,zhang2024dynamic,li20223d,chen2023predicting,li2018learning,shi2023robocook}. While effective in capturing spatial relations, its performance heavily depends on hyperparameters such as adjacency size and connection thresholds, limiting scalability to complex scenes.

    \vspace{-0.3em}
    
    \item \textbf{Scene-centric Modeling:} This paradigm avoids explicit topology construction by directly learning latent dynamics from pixel-based observations. An encoder maps input frames to a unified latent space, where a dynamics model predicts future latent states~\citep{li2022neural,zhou2024dino,hafner2019dream,hafner2023mastering,micheli2022transformers}. However, compressing the entire scene into a single vector often overlooks structural details, impairing generalization in compositional or fine-grained settings.
\end{itemize}

\vspace{-0.3em}

Unlike prior approaches, {\model} learns environment dynamics in an object-centric manner, preserving structural details while addressing the limitations of graph-based methods. It achieves more accurate predictions and removes the need for extensive manual hyperparameter tuning, resulting in more robust and generalizable dynamics modeling.

\vspace{-0.5em}

\paragraph{Multi-Material Robot-Object Interactions.}
Prior works have explored the application of learned world models to a variety of material types, including rigid bodies~\citep{li2019propagation,huang2023defgraspnets,liu2023model}, deformable fabrics~\citep{lin2022learning,longhini2023edo,pfaff2020learning,lin2021softgym,puthuveetil2023robust,huang2023self}, plasticine~\citep{shi2024robocraft,shi2023robocook}, fluid~\citep{sanchez2020learning} and granular materials~\citep{wang2023dynamic,zhang2024dynamic}. However, most of these approaches focus on single-material robot-object scenarios and overlook multi-material interactions, limiting their applicability in real-world settings. In contrast, our work explicitly addresses this gap by extending the benchmark to multi-material robot-object interactions, requiring world models to handle heterogeneous physical properties within a unified framework. Our experiments show that {\model} effectively generalizes in this setting, outperforming existing baselines across diverse material compositions.

\vspace{-0.5em}

\vspace{-0.2em}
\section{Method}
\label{method}
\vspace{-0.5em}

% We begin by describing the problem formulation in Section~\ref{method:formulation}. Section~\ref{method:tbnd} introduces the perception module and the architecture of our Transformer-based neural world model. Section~\ref{method:loss} details the hybrid loss used to provide fine-grained supervision during training. Finally, Section~\ref{method:mpc} introduces the downstream closed-loop model-based motion planning and control procedure built upon the learned world model. The framework of our method is illustrated in Figure~\ref{fig:method}.

\begin{figure}[ht]
\centering
\includegraphics[width=1\linewidth, trim= 0cm 8cm 0cm 0cm, clip]{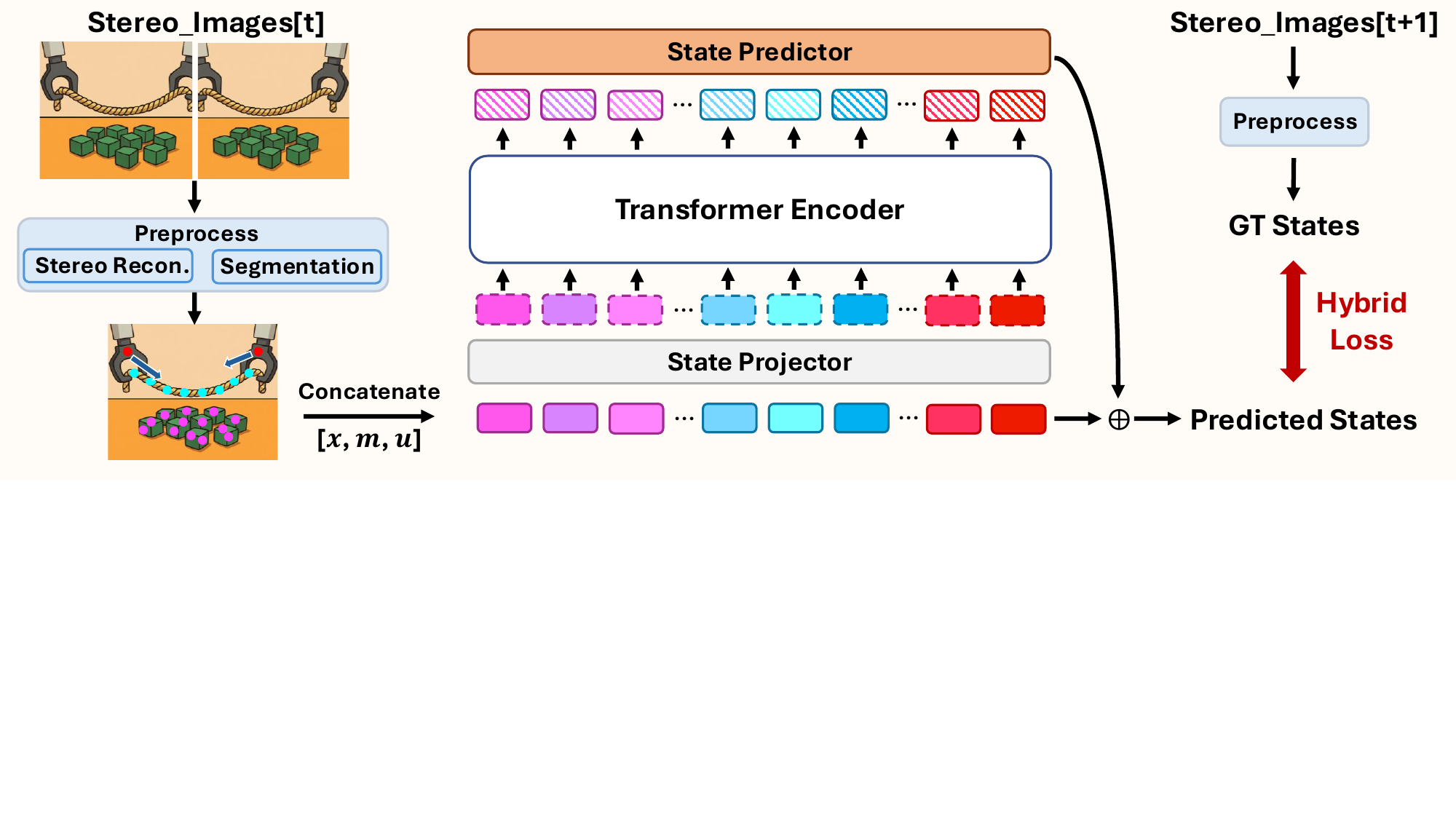}
\caption{\textbf{Overview.} {\model} reconstructs particle-level states from stereo image inputs via stereo matching and segmentation. A Transformer encoder models interaction-aware dynamics over particle features concatenating position, material, and motion cues. The model is trained using a hybrid loss computed against future ground-truth states extracted from the next stereo frames.}
\label{fig:method}
\end{figure}

\subsection{Problem Formulation}
\label{method:formulation}
\vspace{-0.5em}
We consider a robotic manipulation setting where the environment is partially observed at each timestep $t$ as a task-relevant point cloud: $x^{\text{obj}}_t \in \mathbb{R}^{N \times 3}$ for the object state and $x^{\text{ee}}_t \in \mathbb{R}^{M \times 3}$ for the end-effector state, both extracted from the full scene observation. Here, $N$ is the number of downsampled object particles, and $M$ is the number of end-effector points representing its position and shape. Our objective is to learn a neural world model $f$ that predicts the environment's evolution in response to robot actions, with the goal of minimizing the accumulated prediction error over time.

The model takes as input the current environment state $x_t = \{x^{\text{obj}}_t, x^{\text{ee}}_t\}$, the externally applied motion $u_t = \{u^{\text{obj}}_t, u^{\text{ee}}_t\}$, and a material encoding $M = \{m(i)\}$. Specifically, $u^{\text{obj}}_t = 0$ for object particles, and $u^{\text{ee}}_t = x^{\text{ee}}_t - x^{\text{ee}}_{t-1}$ for the end-effector, computed via forward kinematics. The material encoding $M$ contains one-hot vectors, where each $m(i)$ represents the material type of the $i$-th particle (e.g., rigid, granular, rope, or cloth), and $m(i) = 0$ if the $i$-th particle belongs to the end-effector. The world model then predicts the next state as:
\begin{equation}
    x_{t+1} = f(x_t, u_t, M).
    \label{eq:dynamics}
\end{equation}

Conditioning on material type $M$ allows the model to capture material-dependent physical properties and behaviors, which are essential for accurately modeling multi-material interactions.

\subsection{Transformer-Based Neural Dynamics Learning}
\label{method:tbnd}
\vspace{-0.5em}
Our proposed dynamics modeling framework~(as shown in Figure~\ref{fig:method}) is designed to learn particle-level interactions in physical systems using a Transformer-based architecture~\citep{vaswani2017attention} that incorporates multi-head self-attention. At each timestep $t$, the architecture comprises three main components: (i) observation embedding, (ii) dynamics transition, and (iii) motion prediction.

\paragraph{Observation Embedding.} 
\vspace{-0.5em}
This component extracts the state of task-relevant objects and converts pixel-based perception into feature embeddings for downstream dynamics learning. Given a pair of stereo images from the task scene, we first apply FoundationStereo~\citep{wen2025foundationstereo} to reconstruct a dense point cloud. Then, we use GroundingDINO~\citep{liu2024grounding} and Segment Anything~\citep{kirillov2023segment} to segment the specific object of interest being manipulated, yielding a clean point cloud $x^{\text{obj}}_t$ that excludes irrelevant objects. For each particle $i$ at time $t$, we concatenate its position $x_t(i)$, material encoding $m(i)$, and motion $u_t(i)$, and encode the combined feature using an projector $f_{\text{proj}}$ to obtain its latent representation $z_t(i)$:
\begin{equation}
    z_t(i) = f_{\text{proj}}([x_t(i), m(i), u_t(i)]).
    \label{eq:embed}
\end{equation}

\paragraph{Dynamics Transition.} 
\vspace{-0.5em}
At each timestep $t$, we obtain particle-wise embeddings $z_t \in \mathbb{R}^{(N+M) \times d}$, where $d$ is the embedding dimension and $N+M$ denotes the total number of object and end-effector particles. Since the environment is parameterized as a set of particles, learning dynamics reduces to modeling their pairwise interactions. To capture these interactions, we apply a stack of $L = 3$ multi-head self-attention layers, implemented as a dynamics transition module $f_{\text{dyn}}$. The particle embeddings are updated by applying $f_{\text{dyn}}$ to the entire set of latent embeddings:
\begin{equation}
    z'_{t+1} = f_{\text{dyn}}(z_t),
    \label{eq:dyn}
\end{equation}

where $z'_{t+1} \in \mathbb{R}^{(N+M) \times d}$ denotes the predicted latent representation at the next timestep. The module $f_{\text{dyn}}$ consists of multi-head Transformer encoder layers that allow each particle to attend to all others and infer interaction-aware dynamics. Note that $z_t$ already incorporates spatial and motion information, so no additional positional encodings are needed. The attention mechanism directly operates on these embeddings to model physical interactions across particles. In contrast to prior methods that rely on predefined interaction graphs or distance-based adjacency thresholds, our approach learns interaction structures implicitly through attention, reducing sensitivity to such hyperparameters and enabling more flexible modeling in complex, multi-material environments.

\paragraph{Motion Prediction.} 
\vspace{-0.5em}
After the dynamics transition, we obtain updated latent embeddings $z'_{t+1} \in \mathbb{R}^{(N+M) \times d}$, which encode the predicted interaction-aware state at the next timestep. To recover particle-level motion, we apply a shared decoder $f_{\text{pred}}$ to each latent embedding to predict the displacement of each particle:
\begin{equation}
    \Delta \hat{x}_{t+1}(i) = f_{\text{pred}}(z'_{t+1}(i)), \quad \text{for } i = 1, \dots, N+M,
    \label{eq:pred}
\end{equation}
where $\Delta \hat{x}_{t+1}(i) \in \mathbb{R}^3$ denotes the predicted positional change of the $i$-th particle in 3D space.

The predicted next-step particle position is then computed as:
\begin{equation}
    \hat{x}_{t+1}(i) = x_t(i) + \Delta \hat{x}_{t+1}(i).
    \label{eq:update}
\end{equation}

This decoding scheme ensures that the predicted dynamics remain physically grounded in particle-level motion, while maintaining modularity between interaction modeling and spatial state recovery.

\subsection{Hybrid Supervision}
\label{method:loss}
\vspace{-0.5em}
To supervise dynamics learning, we propose a hybrid geometric loss that combines Chamfer Distance~(CD) and a differentiable approximation of Hausdorff Distance~(HD)~\citep{charpiat2005approximations}:
\begin{equation}
    \mathcal{L}_{\text{hybrid}} = \alpha \cdot \mathcal{L}_{\text{CD}} + (1 - \alpha) \cdot \mathcal{L}_{\text{HD}},
    \label{eq:loss}
\end{equation}
where $\alpha \in [0, 1]$ balances fine-grained local alignment and global structural coverage.

Chamfer Distance provides local supervision by minimizing the average nearest-neighbor distance between predicted and ground-truth particles. However, in multi-material scenes with sparse or uneven motion, CD tends to bias the learning toward regions with prominent motion, often neglecting particles affected indirectly through contacts or deformations.

Hausdorff Distance complements this by penalizing large worst-case deviations, helping the model account for subtle but semantically meaningful motions. This hybrid formulation encourages accurate predictions across both dominant and passive regions without requiring point-level correspondences, offering a practical balance between precision and scalability.

\paragraph{Training.} 
\vspace{-0.5em}
We train the world model using the hybrid loss described in Section~\ref{method:loss}, applied over a rollout horizon of $k$ future steps. At each training iteration, we randomly sample a time step $t$ from the dataset as the rollout starting frame, and autoregressively unroll the model for $k$ steps using its own predictions as inputs. In all experiments, we set $k = 5$ to balance long-term supervision and computational efficiency.

\subsection{Model-Based Planning and Control}
\label{method:mpc}
\vspace{-0.5em}
The learned world model $f$ is deployed for downstream robotic manipulation via model-based control. Given the state space $\mathbb{X}$, action space $\mathbb{U}$, and a cost function $c: \mathbb{X} \times \mathbb{X} \times \mathbb{U} \rightarrow \mathbb{R}$, our objective is to optimize a sequence of actions that minimizes the total cost with respect to the current object state $x^{\text{obj}}_t$ and a target state $x^{\text{obj}}_{\text{target}}$. At each control step, we sample action sequences $\{u_i\}_{i=1}^{H}$ over a predefined horizon $H$, roll out predicted trajectories using the world model $f$, and apply the Model Predictive Path Integral~(MPPI) algorithm~\citep{williams2017model} to find the sequence with the lowest expected cost. The cost function includes a task-specific term measuring the distance to the target, as well as penalty terms for infeasible actions and collisions.

\vspace{-0.5em}

\vspace{-0.2em}
\section{Experiment}
\label{exp}
\vspace{-0.5em}

We empirically evaluate {\model} to address the following questions: 
(1) Can it generate visually plausible and physically coherent dynamics, especially in complex multi-material interactions? 
(2) How does its prediction accuracy compare to existing dynamics modeling baselines? 
(3) To what extent does it enhance visuomotor control performance in downstream manipulation tasks?

\subsection{Experiment Setup and Task Design}
\label{exp:setup}
We extend existing dynamics learning benchmarks and evaluate our method on six simulation tasks~(using Nvidia FleX~\citep{macklin2014unified}) and three real-world tasks, with extensive coverage of multi-material, multi-object interactions. Below, we introduce the real-world task designs. Each task involves one or two UFACTORY xArm-6 robot arms and a ZED-2i stereo camera. We collect training data by capturing stereo RGB images and recording end-effector poses at 10 FPS during random robot-object interactions. Details of the experimental setup and benchmarks are provided in Appendix~\ref{app:benchmark}.

\vspace{-0.9em}

\begin{figure}[ht]
\centering
\includegraphics[width=1\linewidth, trim= 0.75cm 0cm 0.72cm 0cm, clip]{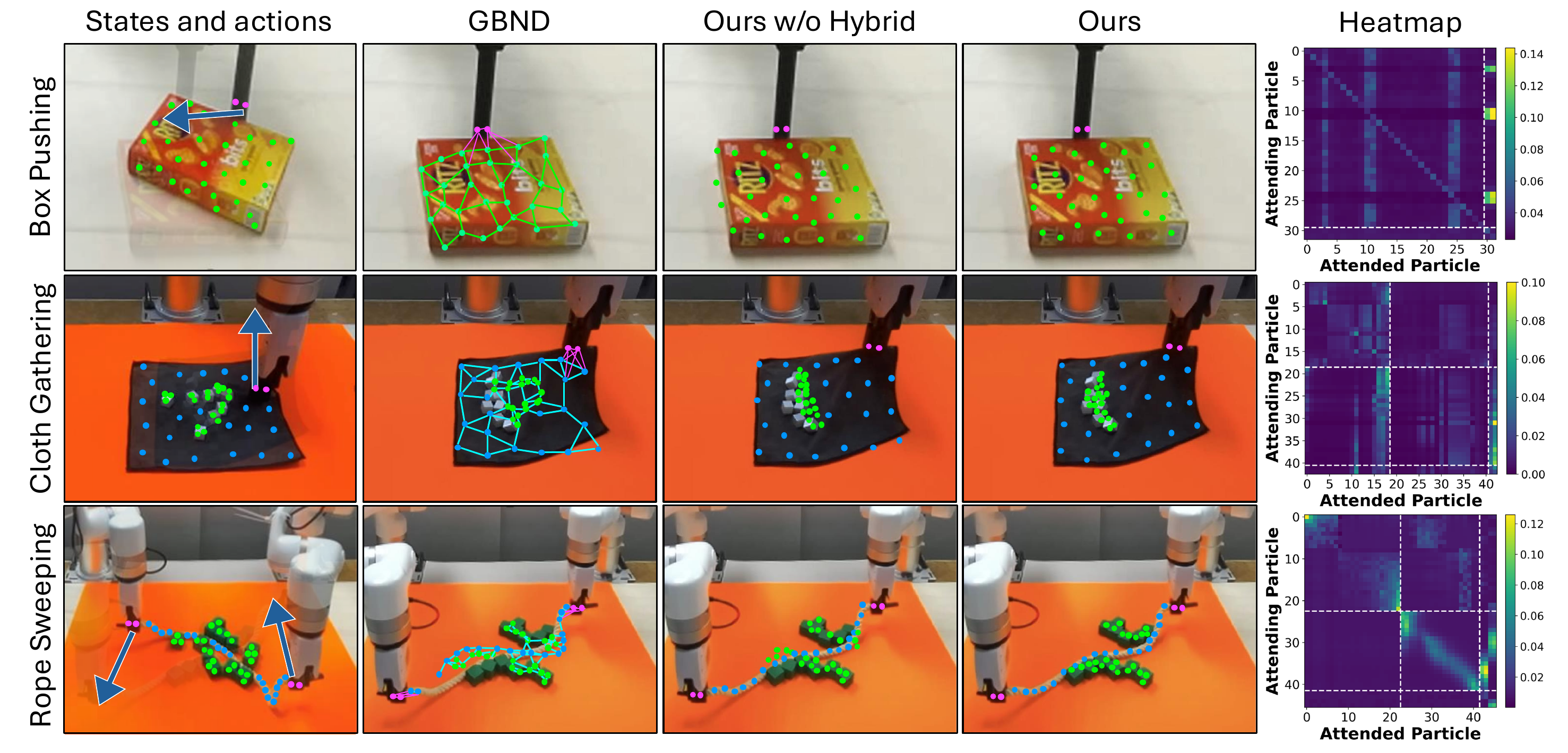}
\vspace{-1em}
\caption{\textbf{Qualitative Results for Dynamics Prediction.} We compare one-step dynamics predictions from {\model} and baseline methods. {\model} demonstrates superior capability in capturing object dynamics and multi-material interactions. The rightmost column shows block-wise attention heatmaps from our method~(attention weights from particle $i$(row) to particle $j$ (column)), revealing the learned interaction structures across both intra- and inter-material particles.}
\label{fig:dyn_qual}
\end{figure}

\vspace{-0.9em}

\paragraph{Box Pushing.}
In this task, the robot end-effector moves horizontally along the table surface, equipped with a cylindrical pusher at its tip. The goal is to push a rigid box continuously toward a desired pose. Successfully completing this task requires {\model} to capture accurate rigid-body dynamics, including both translational and rotational motion, to align the pushing direction with the box’s inertial response.

\vspace{-0.9em}

\paragraph{Cloth Gathering.}
The robot is tasked with gathering granular particles scattered on a cloth. It samples grasping points and vertically lifts the cloth using a two-finger gripper, causing the granular objects to fall and slide due to the induced motion of the fabric. This task demands {\model} to model both the deformable dynamics of the cloth and the passive flow of granular media, enabling accurate prediction of granular motion patterns under localized fabric manipulation.

\vspace{-0.9em}

\paragraph{Rope Sweeping.}
This task involves using two robot arms to sweep scattered granular particles into a designated area with a nylon rope held at both ends. The model is only trained on scenes where the rope is surrounded by granular objects. However, during testing, we place the rope on one side and the granular objects on the other, and ask the model to perform model-based planning and control in this unseen configuration. The robots must coordinate to manipulate the shape and trajectory of the rope to surround and sweep granular material. Success in this task requires {\model} to capture fine-grained rope dynamics and indirect contact-based interactions. % as the rope deforms and transmits force to the granular particles through friction and sweeping motion.

\vspace{-0.9em}

\paragraph{Baselines.}
To demonstrate the effectiveness of {\model}, we compare against three baselines in our main experiments:  
(1) \textit{GBND}, a particle-based GNN world model following~\citet{zhang2024dynamic}, retrained with hybrid loss;  
(2) \textit{Ours w/o Hybrid}, an ablated version of {\model} trained using only Chamfer Distance supervision;  
(3) \textit{DINO-WM}~\citep{zhou2024dino}, a scene-centric world model that builds world models directly from 2D pixel observations using a pretrained vision encoder. % not real-world?

\subsection{Qualitative Evaluation on Dynamics Prediction}
\label{exp:dyn_qualitative}
\vspace{-0.5em}
We first evaluate whether {\model} produces visually accurate and physically plausible dynamics predictions. Figure~\ref{fig:dyn_qual} compares one-step predictions from {\model}, \textit{GBND}, and \textit{Ours w/o Hybrid} across three real-world tasks~(blue arrows indicate the robot’s motion direction). 

\begin{wrapfigure}{r}{0.45\linewidth}
\centering
\vspace{-1em}
\includegraphics[width=\linewidth]{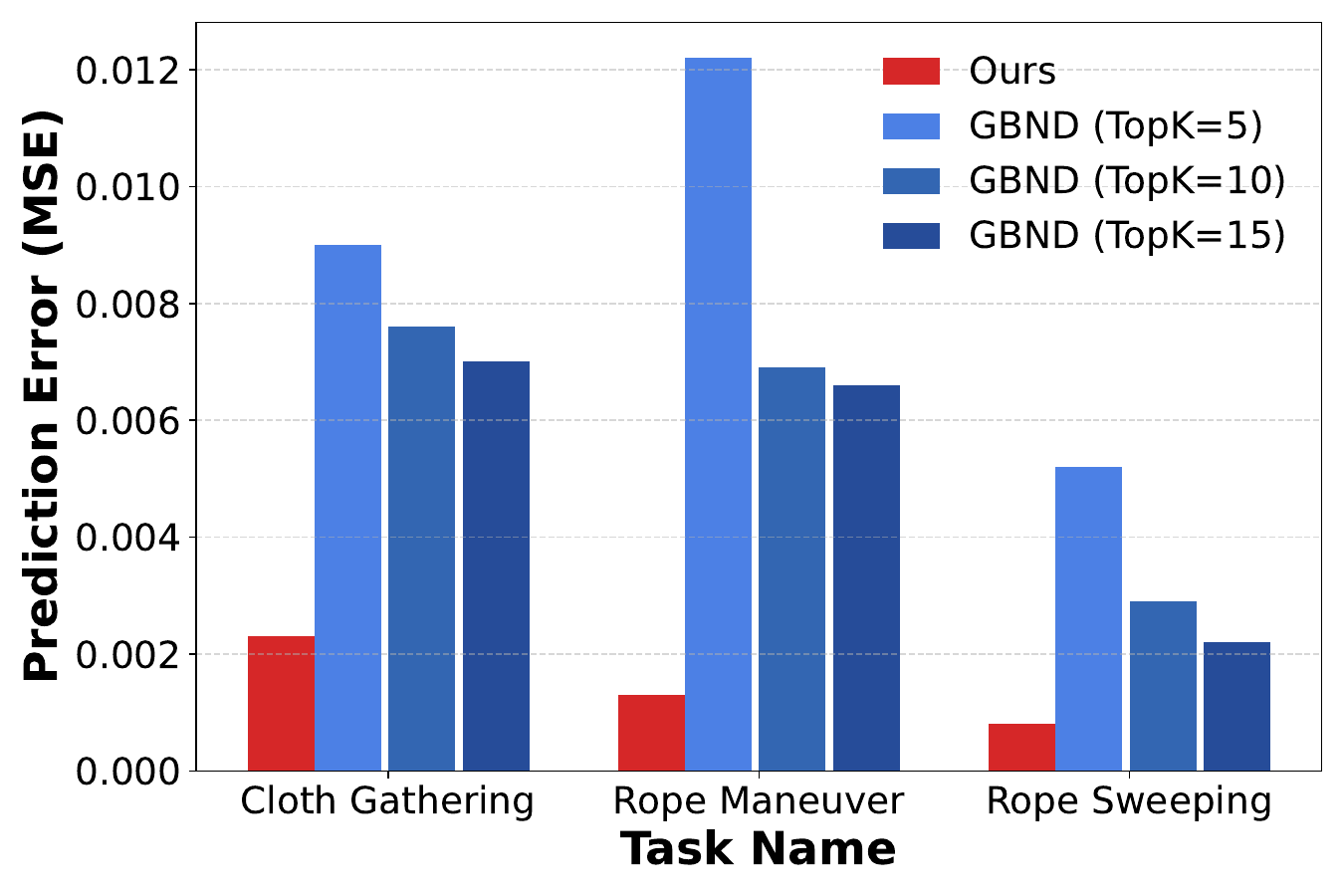}
\caption{\textbf{Effect of \texttt{TopK} on \textit{GBND} Dynamics Accuracy.} Increasing the number of allowed neighbors improves \textit{GBND}'s prediction accuracy but still falls short of {\model}, which achieves lower error without requiring hyperparameter tuning.}
\label{fig:dyn_quan}
\end{wrapfigure}

In the \textbf{Box Pushing} task, {\model} captures both translational and rotational dynamics, while \textit{GBND} produces unrealistic local distortions due to its limited receptive field, and \textit{Ours w/o Hybrid} struggles to align orientation. In the \textbf{Cloth Gathering} task, {\model} accurately models the chain of interaction from cloth lifting to granular flow, whereas \textit{GBND} fails to propagate forces across the cloth, and \textit{Ours w/o Hybrid} underestimates the resulting granular movement. In the \textbf{Rope Sweeping} task, {\model} correctly predicts both rope deformation and induced granular displacement, while \textit{GBND} fails to propagate motion through the rope, resulting in a split rope graph and unrealistic prediction; \textit{Ours w/o Hybrid} again underpredicts the motion of indirectly affected granular media, underscoring the value of hybrid supervision. 

The final column shows attention heatmaps from {\model}. For instance, in the Rope Sweeping task, distinct structures emerge: rope-rope interactions form a narrow linear band along the diagonal, consistent with the rope's sequential connectivity; granular-granular interactions appear as two broader blocks, reflecting spatially separated granular clusters; strong granular-to-rope attention highlights the force transfer from the rope to the granular particles; and rope particles attend asymmetrically to different robot end-effectors at either end. These patterns indicate that our method captures fine-grained material interactions directly from data without predefined topologies.

% The final column shows attention heatmaps from {\model}, where block structures along the diagonal indicate interactions within the same material category (e.g., cloth-cloth or granular-granular), while off-diagonal blocks reveal interactions across different material types and robots (e.g., robot-box, rope-granular). For instance, in the \textbf{Rope Sweeping} task, ... These patterns demonstrate that {\model} automatically learns to attend to physically relevant particles both within and across materials, effectively modeling material-aware interactions without requiring predefined topologies or interaction hyperparameters.

\begin{table*}[t]
\centering
\caption{\textbf{Quantitative Results for Dynamics Prediction.} We report prediction errors across three multi-material simulation tasks~(please refer to Appendix~\ref{app:benchmark} for detailed task descriptions), where ground-truth object state information is available for computing evaluation metrics. Since \textit{DINO-WM} does not provide point-to-point particle correspondences, we evaluate it using point cloud-based metrics only. Results for additional tasks are provided in Appendix~\ref{app:dyn}.}
\label{tab:dyn_quan}
\begin{tabular}{ll|ccc}
\toprule
\textbf{Metric} & \textbf{Method} 
& \textbf{Cloth Gathering}
& \textbf{Rope Maneuver}
& \textbf{Rope Sweeping} \\
\midrule

\multirow{2}{*}{MSE~$\downarrow$} 
& Ours                             & \textbf{0.0023} & \textbf{0.0013} & \textbf{0.0008} \\
& Ours w/o Hybrid                  & 0.0042          & 0.0037          & 0.0013 \\
& GBND~\citep{zhang2024dynamic}    & 0.0076          & 0.0069          & 0.0029 \\
\midrule

\multirow{3}{*}{CD~$\downarrow$} 
& Ours                             & 0.096          & \textbf{0.071} & 0.067 \\
& Ours w/o Hybrid                  & \textbf{0.082} & 0.074          & \textbf{0.061} \\
& GBND~\citep{zhang2024dynamic}    & 0.152          & 0.183          & 0.107 \\
& DINO-WM~\citep{zhou2024dino}     & 0.187          & 0.233          & 0.126 \\
\midrule

\multirow{3}{*}{CD+HD~$\downarrow$} 
& Ours                             & \textbf{0.484} & \textbf{0.216} & \textbf{0.328} \\
& Ours w/o Hybrid                  & 0.749          & 0.673          & 0.415 \\
& GBND~\citep{zhang2024dynamic}    & 0.886          & 0.628          & 0.473 \\
& DINO-WM~\citep{zhou2024dino}     & 1.053          & 0.937          & 0.798 \\

\bottomrule
\end{tabular}
\end{table*}

\subsection{Quantitative Evaluation on Dynamics Prediction}
\label{exp:dyn_quantitative}
\vspace{-0.5em}
The quantitative results in Table~\ref{tab:dyn_quan} show that {\model} consistently achieves the best performance across all metrics, particularly in MSE and CD+HD, indicating superior accuracy in both local details and global structure. To enable a fair comparison with \textit{DINO-WM}, which produces only 2D scene predictions rather than 3D particle states, we additionally use Depth Anything V2~\citep{yang2024depth} to reconstruct particle point clouds from its predicted images in order to compute 3D metrics.

% The quantitative results in Table~\ref{tab:dyn_quan} show that {\model} consistently achieves the best performance across all metrics, particularly in MSE and CD+HD, indicating superior accuracy in both local details and global structure. While \textit{Ours w/o Hybrid}, trained solely with Chamfer Distance~(CD), achieves slightly lower CD scores in some tasks, it performs significantly worse on other metrics—suggesting that optimizing only CD leads to imbalanced learning. In contrast, the hybrid loss used in {\model} offers more comprehensive supervision, resulting in more reliable and generalizable dynamics prediction.

Additionally, \textit{GBND} exhibits strong sensitivity to hyperparameter settings. Specifically, we evaluate its dynamics prediction performance under varying values of a crucial variable: the maximum number of allowed adjacent nodes~(i.e., \texttt{TopK}), as shown in Figure~\ref{fig:dyn_quan}. As this parameter increases from 5 to 15, the prediction error decreases significantly. However, the computational cost also increase rapidly~(Appendix~\ref{app:dyn}), due to the need to store and process larger adjacency matrices—making it impractical to increase this value indefinitely. In contrast, {\model} leverages the soft-attention mechanism through its Transformer-based architecture, which eliminates the need for manually tuning such graph-related hyperparameters. It achieves lower prediction error with less cost, offering benefits in both performance and scalability.

\subsection{Model-Based Planning and Control}
\label{exp:mpc}
\begin{figure}[ht]
\centering
\includegraphics[width=0.95\linewidth]{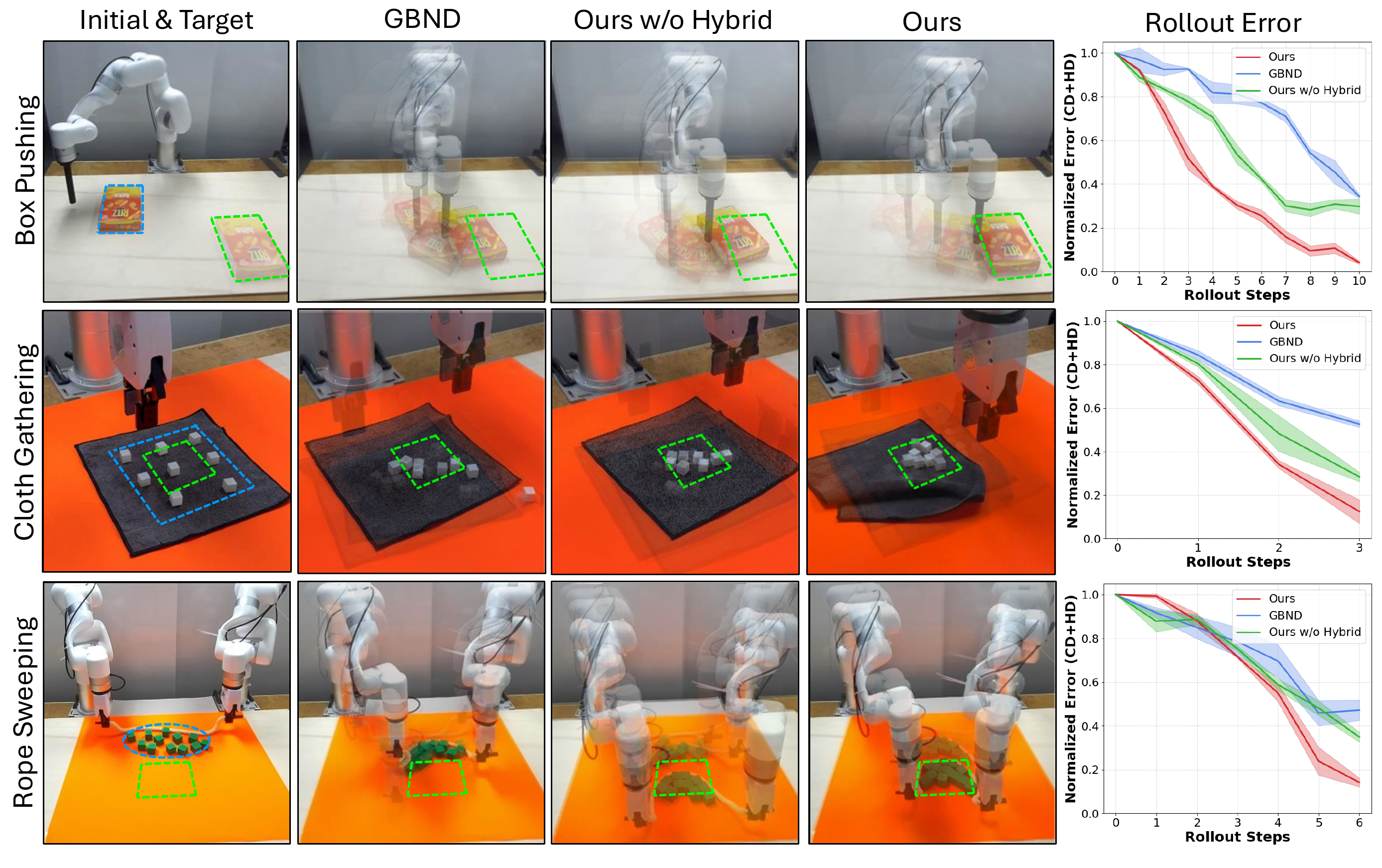}
\vspace{-1em}
\caption{\textbf{Experimental Results for MPC Rollout.} The robot is tasked with using the learned world model to perform closed-loop feedback control toward novel target states unseen during training. The initial state is indicated by blue dashed boxes, and the target state is indicated by green dashed boxes. Compared to baselines, {\model} achieves more accurate planning and control, exhibiting the lowest final-state mismatch over three rollout trials.}
\label{fig:mpc}
\end{figure}

\vspace{-0.5em}
Finally, we evaluate whether the learned world model enables effective downstream robot manipulation via model-based control. We deploy {\model} with the MPPI~\citep{williams2017model} algorithm to achieve goal-directed behaviors toward novel target configurations unseen during training, using the normalized CD+HD as the cost function for planning. Figure~\ref{fig:mpc} presents both qualitative and quantitative results across the three real-world tasks introduced in Section~\ref{exp:setup}. Compared to baselines, {\model} demonstrates superior capability in leveraging accurate dynamics prediction, consistently achieving lower final-state errors and producing more fine-grained manipulation behaviors, even in complex multi-material scenarios. Please refer to Appendix~\ref{app:mpc} for more results.

\vspace{-0.5em}

\vspace{-0.9em}
\section{Conclusion}
\vspace{-0.9em}

In this paper, we present {\model}, a state-of-the-art world model that achieves superior performance in challenging robotic tasks. {\model} improves dynamics modeling accuracy and robustness through two key designs: a Transformer-based particle-parametrization framework that flexibly models soft interaction dependencies among environmental particles, and a hybrid point cloud supervision loss that captures both global structure and local motion details. Furthermore, we extend evaluation benchmarks from single-material to multi-material scenarios and demonstrate the effectiveness of {\model} across both synthetic and real-world settings. We believe {\model} provides a capable and easy-to-use dynamics modeling framework, with potential to push the boundaries of world model applications in robotic manipulation tasks.

\clearpage

\paragraph{Limitations.} While {\model} has demonstrated its efficacy, it is currently trained per scene and has not yet been shown to generalize across a wide variety of environments and robots. In addition, the pipeline depends on externally generated object masks obtained with off-the-shelf foundation models; segmentation failures can propagate to the dynamics prediction. Future work could explore training a single, scene-agnostic model on broader 3D robotic datasets and embedding semantic codes directly into the world model without explicit masking.

\paragraph{Acknowledgment.} This work was supported in part by the U.S. National Science Foundation under Grant No. FRR-2342246.

\bibliography{main}  % .bib

\clearpage
\appendix
\section*{Appendix}
\section{Experiment Implementation Details}
\label{app:benchmark}

\subsection{Setup}
Our experiments focus on robotic manipulation tasks where {\model} must model the dynamics of diverse objects, including complex multi-material interactions. The tasks span both single-material and multi-material robot-object interactions, and are used to evaluate dynamics modeling accuracy as well as downstream visuomotor control performance. We run all dynamics learning experiments on an NVIDIA A100 GPU and perform real-world model-based visuomotor control on an NVIDIA RTX 3090 GPU.

\begin{figure}[ht]
\centering
\includegraphics[width=1\linewidth, trim= 5cm 6.5cm 5cm 0cm, clip]{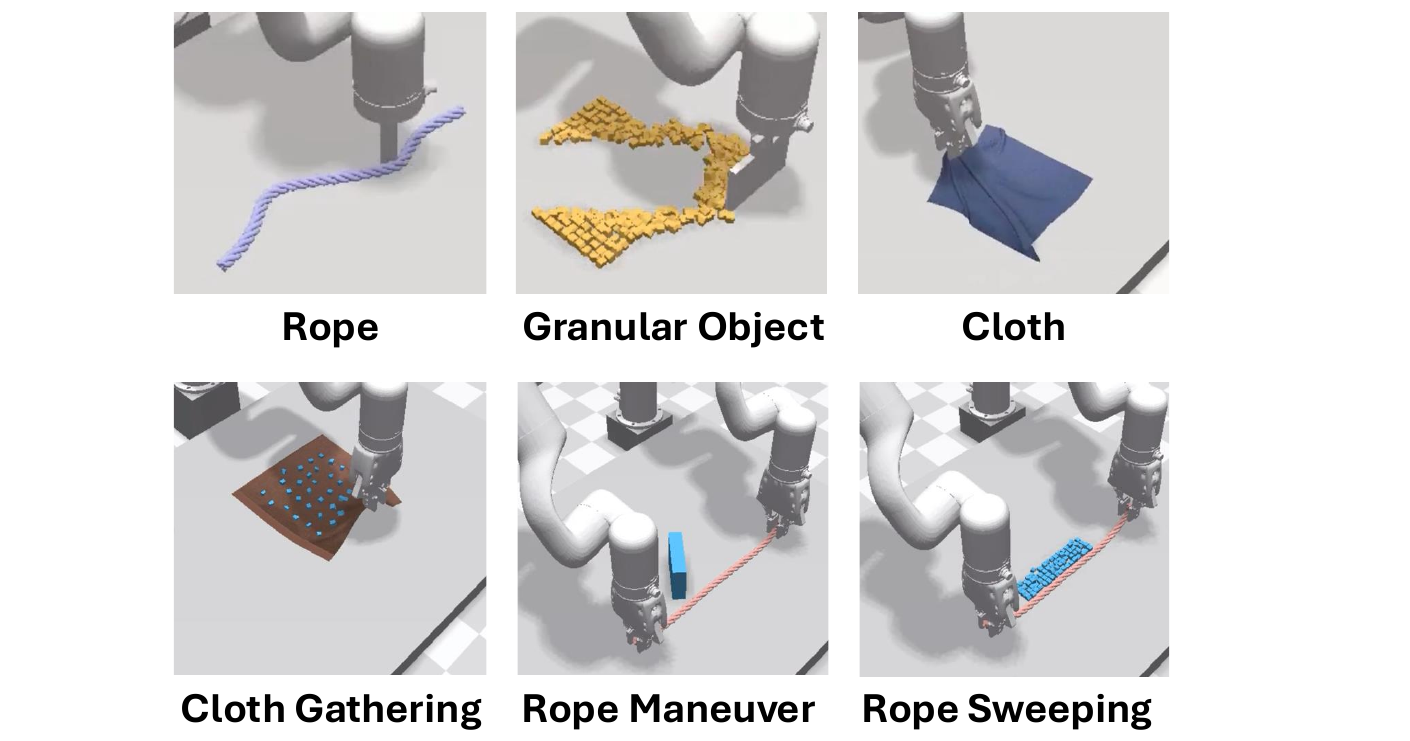}
\caption{\textbf{Simulation Experiment Setup.} }
\label{fig:app_sim_setup}
\end{figure}

\paragraph{Simulation.} As shown in Figure~\ref{fig:app_sim_setup}, we extend the existing dynamics modeling benchmark created by~\citet{zhang2024adaptigraph} based on the NVIDIA FleX simulator~\citep{macklin2014unified}. The original tasks, which involve only single-material robot-object interactions—Rope, Granular Object, and Cloth—are used solely for evaluating dynamics modeling performance. We additionally design three tasks involving multi-material robot-object interactions, used to evaluate both dynamics modeling and downstream visuomotor control. All tasks use the UFACTORY xArm-6 robot, and the end-effectors include a cylindrical pusher, a planar pusher, and a parallel gripper. We collect 1,000 episodes of data for each task as the training dataset.

\begin{figure}[ht]
\centering
\includegraphics[width=1\linewidth, trim= 5cm 3cm 5cm 0cm, clip]{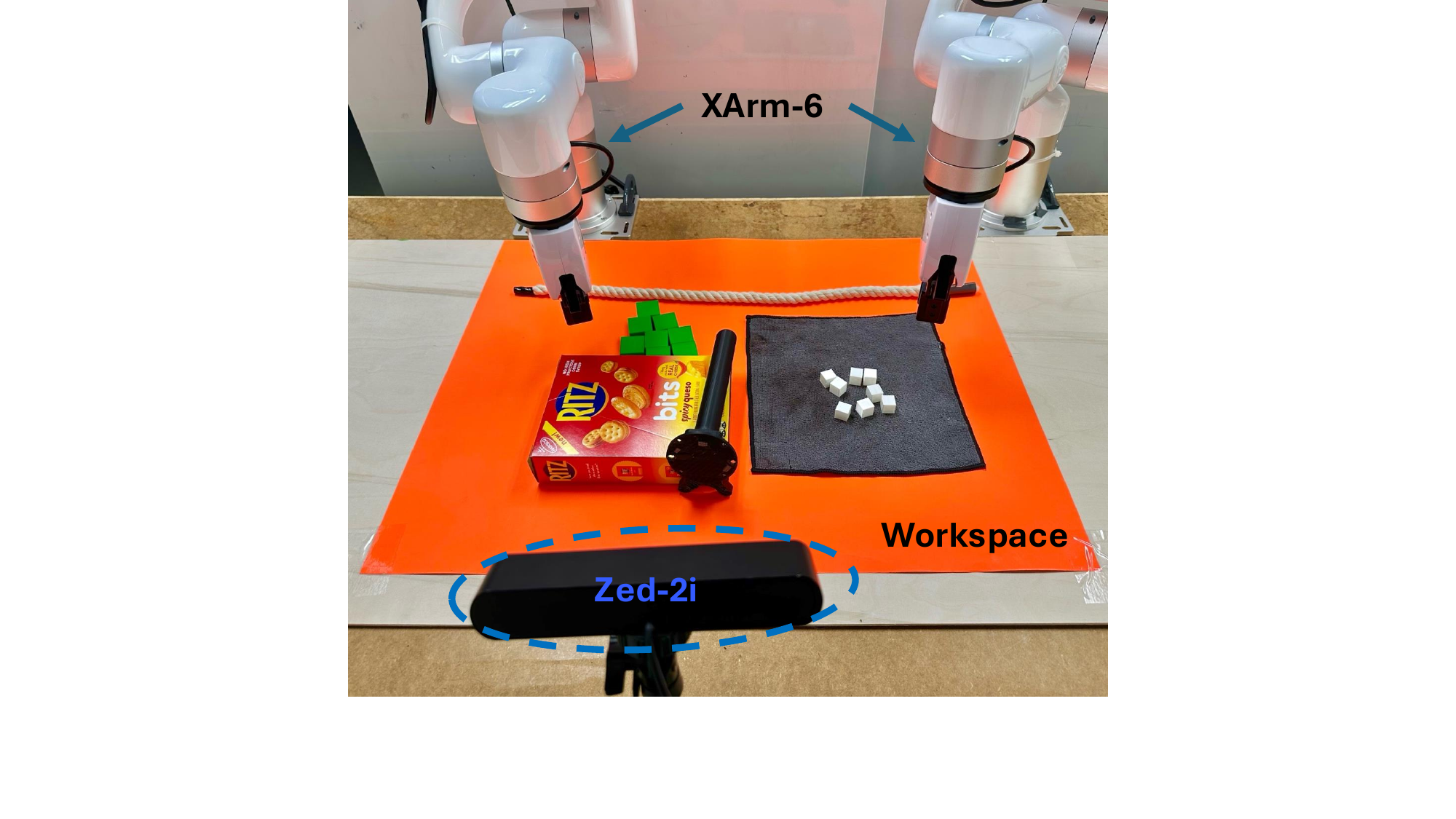}
\caption{\textbf{Real-World Experiment Setup.}}
\label{fig:app_real_setup}
\end{figure}

\paragraph{Real World.} In our real-world experiments, as shown in Figure~\ref{fig:app_real_setup}, we utilize one UFACTORY xArm-6 robotic arm (for the Box Pushing and Cloth Gathering tasks) or two arms (for the Rope Sweeping task), each with 6 DoF. A 3D-printed cylindrical pusher is attached for the Box Pushing task, while a parallel gripper is used for the Cloth Gathering task and two grippers for the Rope Sweeping task. A ZED-2i camera is positioned in front of the workspace to capture stereo RGB images at 10 Hz with a resolution of 1280×720. The robot manipulation workspace measures approximately 70 cm × 55 cm. We record 500-second episodes of random robot-object interactions for each task to construct the training dataset.

\subsection{Task Design}
We introduce the tasks used to evaluate the performance of {\model} compared to baselines, spanning both single-material and multi-material interactions.

\paragraph{Rope.}
The robot uses a cylindrical pusher to randomly push a rope placed in the planar workspace. This task evaluates whether the world model can accurately capture the deformable dynamics of the rope and its interactions with the robot.

\paragraph{Granular Object.}
The robot uses a flat pusher to randomly push piles of granular material scattered across the workspace. This task tests the model's ability to predict the dynamics of disjoint granular particles under direct physical contact.

\paragraph{Cloth.}
The robot randomly grasps the corner of a cloth using a parallel gripper and moves it along different trajectories. The goal is to evaluate the model’s ability to capture the deformable dynamics of the fabric under manipulation.

\paragraph{Box Pushing.}
In this task, the robot end-effector moves horizontally along the table surface, equipped with a cylindrical pusher at its tip. The goal is to push a rigid box continuously toward a desired pose. Successfully completing this task requires {\model} to capture accurate rigid-body dynamics, including both translational and rotational motion, to align the pushing direction with the box’s inertial response.

\paragraph{Cloth Gathering.}
The robot is tasked with gathering granular particles scattered on a cloth. It samples grasping points and vertically lifts the cloth using a two-finger gripper, causing the granular objects to fall and slide due to the induced motion of the fabric. This task demands {\model} to model both the deformable dynamics of the cloth and the passive flow of granular media, enabling accurate prediction of granular motion patterns under localized fabric manipulation.

\paragraph{Rope Maneuver.}
In this task, two robot arms grasp and manipulate the ends of a rope to indirectly move a rigid box placed on the ground. The objective is to maneuver the box to a target pose without tipping it over. This setting features rich multi-object, multi-material interactions and requires {\model} to reason about indirect force transmission between deformable and rigid bodies.

\paragraph{Rope Sweeping.}
This task involves using two robot arms to sweep scattered granular particles into a designated area with a nylon rope held at both ends. The robots must coordinate to manipulate the shape and trajectory of the rope to surround and sweep granular material. Success in this task requires {\model} to capture fine-grained rope dynamics and indirect contact-based interactions, as the rope deforms and transmits force to the granular particles through friction and sweeping motion.

In this paper, we evaluate {\model} on six simulation tasks—\textbf{Rope}, \textbf{Granular Object}, \textbf{Cloth}, \textbf{Cloth Gathering~(Sim)}, \textbf{Rope Maneuver}, and \textbf{Rope Sweeping~(Sim)}—and three real-world tasks—\textbf{Box Pushing}, \textbf{Cloth Gathering~(Real)}, and \textbf{Rope Sweeping~(Real)}—to extensively assess its performance in dynamics modeling under complex scenarios, in comparison with leading baselines.

\section{Additional Results for Dynamics Prediction}
\label{app:dyn}
This section provides additional results for dynamics modeling comparison in the simulation.

\subsection{Dynamics Prediction Comparison}
Table~\ref{tab:app_dyn_quan} presents the dynamics prediction performance comparison on three single-material tasks. The results demonstrate the superior performance of {\model} in terms of both MSE and CD+HD metrics, highlighting its ability to model dynamics accurately at both the global structural level and in fine-grained local movements. Results for multi-material tasks are provided in Table~\ref{tab:dyn_quan}.

\begin{table*}[h]
\centering
\caption{\textbf{Addition Quantitative Results for Dynamics Prediction.} We report prediction errors across the three single-material simulation tasks, where ground-truth object state information is available for computing evaluation metrics. Since \textit{DINO-WM} does not provide point-to-point particle correspondences, we evaluate it using point cloud-based metrics only.}
\label{tab:app_dyn_quan}
\begin{tabular}{ll|ccc}
\toprule
\textbf{Metric} & \textbf{Method} 
& \textbf{Rope}
& \textbf{Granular Object}
& \textbf{Cloth} \\
\midrule

\multirow{2}{*}{MSE~$\downarrow$} 
& Ours                             & \textbf{0.00017} & \textbf{0.00152} & \textbf{0.00160} \\
& Ours w/o Hybrid                  & 0.00025          & 0.00173          & 0.00182 \\
& GBND~\citep{zhang2024dynamic}    & 0.00043          & 0.00196          & 0.00194 \\
\midrule

\multirow{3}{*}{CD~$\downarrow$} 
& Ours                             & 0.016          & 0.062          & 0.063 \\
& Ours w/o Hybrid                  & \textbf{0.013} & \textbf{0.056} & \textbf{0.059} \\
& GBND~\citep{zhang2024dynamic}    & 0.045          & 0.074          & 0.076 \\
& DINO-WM~\citep{zhou2024dino}     & 0.052          & 0.125          & 0.130 \\
\midrule

\multirow{3}{*}{CD+HD~$\downarrow$} 
& Ours                             & \textbf{0.149} & \textbf{0.413} & \textbf{0.488} \\
& Ours w/o Hybrid                  & 0.178          & 0.433          & 0.527 \\
& GBND~\citep{zhang2024dynamic}    & 0.291          & 0.470          & 0.664 \\
& DINO-WM~\citep{zhou2024dino}     & 0.304          & 0.527          & 1.425 \\

\bottomrule
\end{tabular}
\end{table*}

\subsection{Hyperparameter Analysis}
\begin{figure}[ht]
\centering
\includegraphics[width=0.5\linewidth]{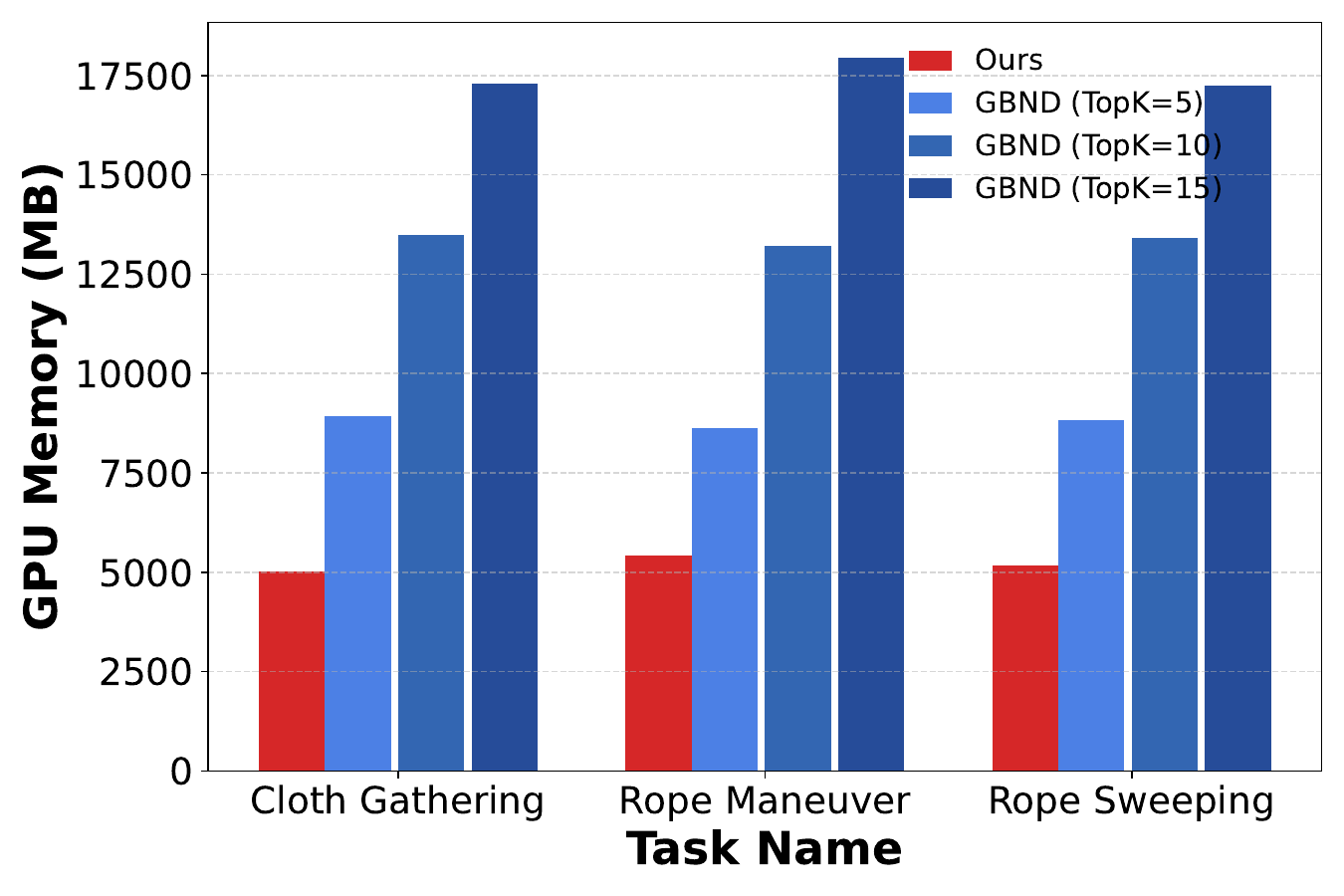}
\caption{\textbf{Effect of \texttt{TopK} on \textit{GBND} GPU Usage.} As the maximum number of allowed adjacent nodes increases, \textit{GBND}'s GPU memory usage grows significantly. This highlights the scalability bottleneck in GNN-based methods. In contrast, {\model} avoids this issue by using soft attention without explicit neighbor selection.}
\label{fig:app_dyn_quan_m}
\end{figure}

As a complementary result to Figure~\ref{fig:dyn_quan}, we analyze the effect of the \texttt{TopK} hyperparameter in \textit{GBND}~\citep{zhang2024dynamic} on GPU memory usage during training~(All models are evaluated using input clusters of 200 scene particles). Figure~\ref{fig:app_dyn_quan_m} shows that increasing \texttt{TopK} leads to significantly higher memory consumption per epoch, reflecting the scalability challenges of graph-based dynamics models.

\begin{figure}[ht]
\centering
\includegraphics[width=0.5\linewidth]{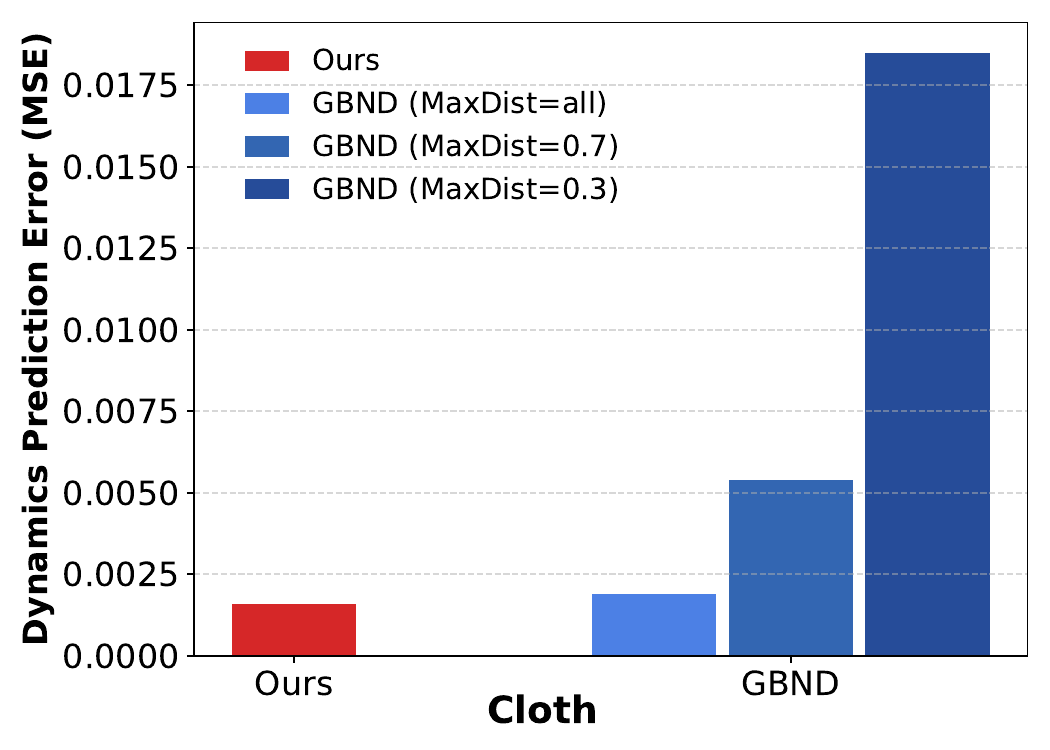}
\caption{\textbf{Effect of \texttt{MaxDist} on \textit{GBND} Dynamics Accuracy.} Since information propagation in \textit{GBND} relies on graph topology, its performance is highly sensitive to the maximum distance threshold for edge construction. In contrast, {\model} avoids this sensitivity through attention-based interactions.}
\label{fig:app_dyn_quan_s}
\end{figure}

Additionally, we observe that \textit{GBND} exhibits strong sensitivity to another key hyperparameter: the maximum distance threshold used to construct edges between particles, denoted as \texttt{MaxDist}. This parameter directly affects the range over which information can propagate in a single message-passing step within the graph. We conduct ablation experiments on the Cloth task to analyze its impact, as shown in Figure~\ref{fig:app_dyn_quan_s}. Specifically, we compare the dynamics prediction error (MSE) under three conditions: (1) when the end-effector particle is connected to all cloth particles, (2) when only short-range connections are allowed (\texttt{MaxDist} = 0.3), and (3) when longer-range connections are allowed (\texttt{MaxDist} = 0.7). The results reveal that the dynamics accuracy of \textit{GBND} varies significantly depending on this setting, highlighting its vulnerability to edge construction heuristics. In contrast, {\model} leverages a Transformer encoder with soft attention to bypass these limitations, achieving more stable and accurate performance.

\section{Additional Results for Model-Based Planning and Control}
\label{app:mpc}

\begin{figure}[ht]
\centering
\includegraphics[width=1\linewidth]{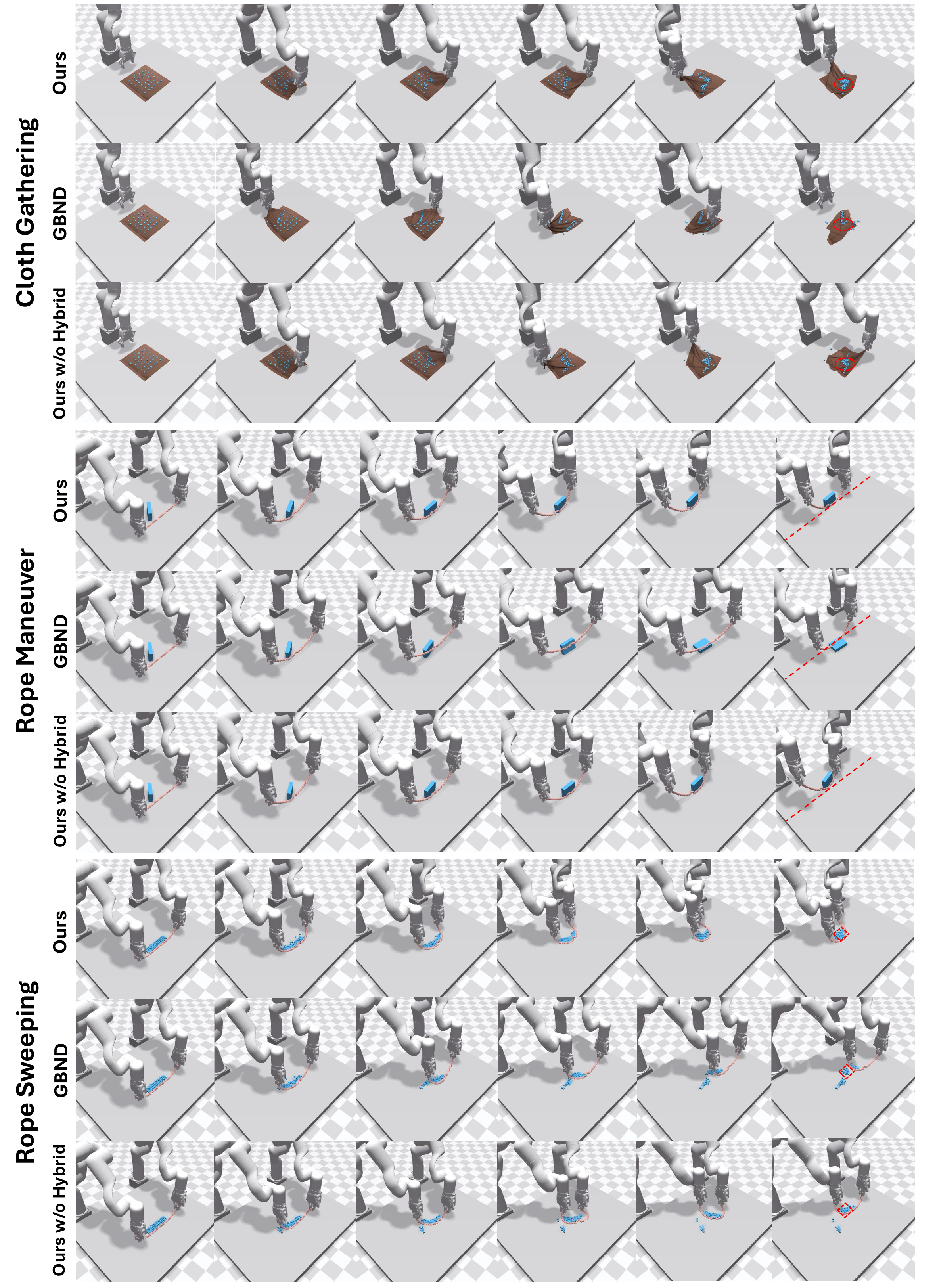}
\caption{\textbf{MPC Rollout Results in Multi-Material Simulation Tasks.} The robot is tasked with using the learned world model to perform closed-loop feedback control toward novel target states unseen during training. Target states are indicated by red dashed boxes or lines. Compared to baselines, {\model} achieves more accurate planning and lower final-state mismatch.}
\label{fig:app_mpc_sim}
\end{figure}

In this section, we provide additional results on model-based visuomotor control using world models trained with {\model} and baseline methods across three multi-object, multi-material simulation tasks. The qualitative results are presented in Figure~\ref{fig:app_mpc_sim}, demonstrating that {\model} enables more fine-grained and higher-quality control behavior in downstream manipulation tasks compared to prior approaches.

\end{document}